\documentclass{article}
\usepackage{spconf,amsmath}

\input{math_commands.sty}
\usepackage{booktabs}
\usepackage[ruled]{algorithm}
\usepackage{algpseudocode}
\title{Fast and Exact Enumeration of Deep Networks Partitions Regions}
\twoauthors
 {Randall Balestriero}
	{Meta AI, FAIR\\
	\texttt{rbalestriero@meta.com}\\
	NYC, USA}
 {Yann LeCun}
	{Meta AI, FAIR, NYU\\
	\texttt{yann@meta.com}\\
	NYC, USA}
\begin{document}
%\ninept
%
\maketitle
\begin{abstract}
One fruitful formulation of Deep Networks (DNs) enabling their theoretical study and providing practical guidelines to practitioners relies on Piecewise Affine Splines. In that realm, a DN's input-mapping is expressed as per-region affine mapping where those regions are implicitly determined by the model's architecture and form a partition of their input space. That partition --which is involved in all the results spanned from this line of research-- has so far only been computed on $2/3$-dimensional slices of the DN's input space or estimated by random sampling. In this paper, we provide the first parallel algorithm that does exact enumeration of the DN's partition regions. The proposed algorithm enables one to  finally assess the closeness of the commonly employed approximations methods, e.g. based on random sampling of the DN input space. One of our key finding is that if one is only interested in regions with ``large'' volume, then uniform sampling of the space is highly efficient, but that if one is also interested in discovering the ``small'' regions of the partition, then uniform sampling is exponentially costly with the DN's input space dimension. On the other hand, our proposed method has complexity scaling linearly with input dimension and the number of regions.
\end{abstract}

\vspace{-0.1cm}
\section{Introduction}
\vspace{-0.1cm}

\begin{figure}[t!]
    \centering
    \includegraphics[width=\linewidth]{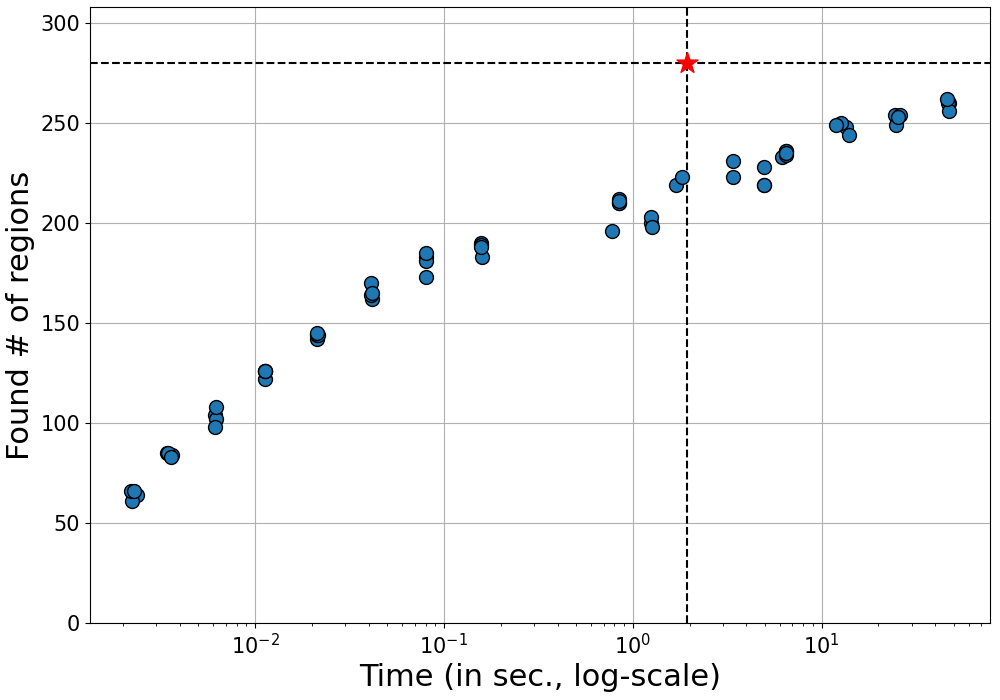}\\
    \includegraphics[width=\linewidth]{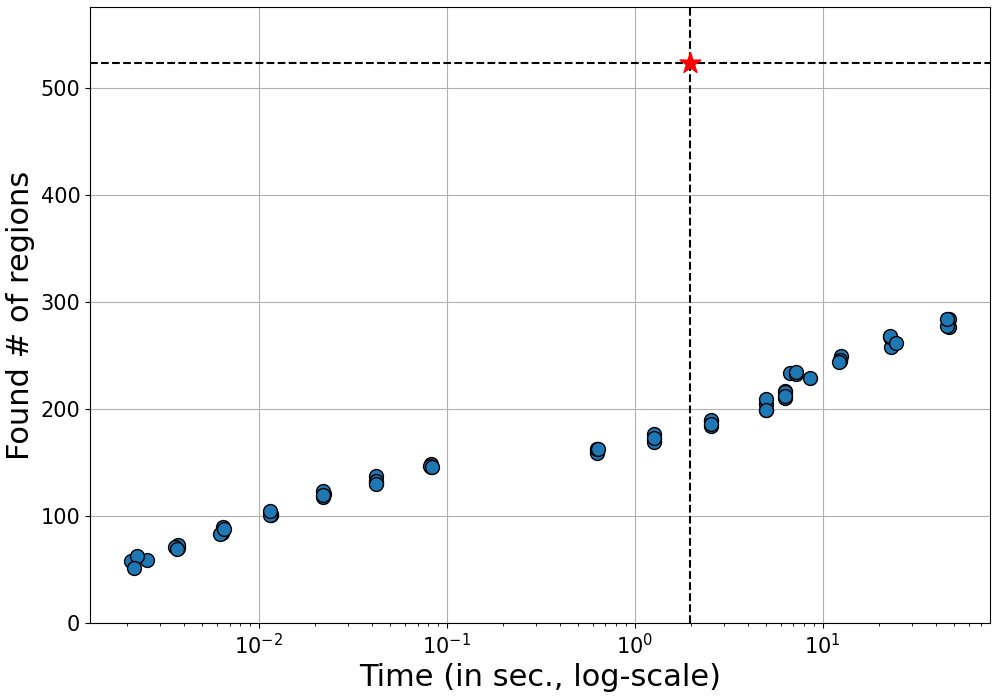}\\
    \vspace{-0.6cm}
    \caption{\small Proposed exact region enumeration depicted as an \color{orange}{\bf orange star} \color{black} against sampling-based region discovery of the partition $\Omega$ depicted as \color{blue}{\bf blue dots} \color{black} for a single hidden layer DN with leaky-ReLU, random parameters and width $64$ as a function of computation time ({\bf x-axis}) and number of partition regions found ({\bf y-axis}); for a $4$-dimensional input space at the {\bf top} and $8$-dimensional input space at the {\bf bottom}. {\em The proposed \cref{algo:police} is able to dramatically outperform the sampling-based search that has been used throughout recent studies on CPA DNs.}}
    \vspace{-0.4cm}
    \label{fig:teaser}
\end{figure}

Deep Networks (DNs) are compositions of linear and nonlinear operators altogether forming a differentiable functional $f_{\vtheta}$ governed by some trainable parameters $\vtheta$ \cite{lecun2015deep}. Understanding the underlying properties that make DNs the great function approximators that they are involve many different research directions e.g. the underlying implicit regularization of architectures \cite{neyshabur2014search}, or the impact of input and feature normalization into the optimization landscape \cite{le1991eigenvalues}. Most existing results emerge from a few different mathematical formulations. One eponymous example relies on kernels and emerges from pushing the DN's layers width to infinity. In this case, and under some additional assumptions, a closed-form expression for the DN's underlying embedding space metric is obtained \cite{jacot2018neural}. With that form, training dynamics and generalization bounds are amenable to theoretical analysis \cite{huang2020deep}. Another line of research considers the case of deep linear networks i.e. a DN without nonlinearities. In this setting, it is possible to obtain the explicit regularizer that acts upon the DN's functional and that depends on the specifics of the architecture e.g. depth and with \cite{soudry2018implicit}. Another direction, most relevant to our study, proposes to unravel the Continuous Piecewise Affine (CPA) mapping of standard DNs \cite{balestriero2018spline}. In short, one can combine the fact that (i) the nonlinearities present in most current DNs are themselves CPA e.g. (leaky-)ReLU, absolute value, max-pooling, (ii) the interleaved affine mappings preserve the CPA property, and (iii) composition of CPA mappings remain CPA. Thus, the entire input-output DN is itself a CPA. From that observation, it is possible to study the DN's loss landscape \cite{riedi2022singular}, the implicit regularizer of different architectures \cite{balestriero2018hard}, the explicit probabilistic distributions of CPA Deep Generative Networks \cite{balestriero2020analytical,humayun2022polarity}, the approximation rates \cite{daubechies2022nonlinear,balestriero2022batch}, or even the conditions for adversarial robustness guarantees \cite{weng2018towards,raghunathan2018semidefinite}. A striking benefit of the CPA viewpoint lies in the fact that it is an exact mathematical description of the DN's input-output mapping without any approximation nor simplification. This makes the obtained insights and guidelines highly relevant to improve currently deployed state-of-the-art architectures.

Despite this active recent development of CPA-based results around DNs, one key challenge remains open. In fact, because under this view one expresses the DN mapping as a collection of affine mappings --one for each region $\omega$ of some partition $\Omega$ of their input space-- it becomes crucial to compute that partition $\Omega$ or at least infer some statistics from it. Current analytical characterizations of $\Omega$ are in fact insufficient e.g. existing bounds characterizing the number of regions in $\Omega$ are known to be loose and uninformative \cite{edelsbrunner1987algorithms}. As such, practitioners resort to simple approximation strategies, e.g. sampling, to estimate such properties of $\Omega$. Another approach is to only consider $2/3$-dimensional slices of the DN's input space and estimate $\Omega$ restricted on that subspace. All in all, nothing is known yet about how accurate are those approximations at conveying the underlying properties of the entire partition $\Omega$ that current theoretical results heavily rely on. In particular, \cite{chen2021neural} uses estimates of the partition's number of region to perform Neural Architecture Search (NAS), and for which exact computation of the DNN's partition regions will further improve the NAS; \cite{humayun2022polarity} uses estimates of the partition to adapt the distribution of deep generative networks (e.g. variational autoencoders) and for which exact computation of the partition would make their method exact, and not an approximation

In this paper, we propose a principled and provable enumeration method for DNs partitions (\cref{algo:police}) that we first develop for a layer-wise analysis in \cref{sec:layer} and then extend to the multilayer case in \cref{sec:deep}. As depicted in \cref{fig:teaser}, the proposed method becomes exponentially faster than the sampling-based strategy to discover the regions $\omega \in \Omega$ as the input dimensionality increases. Practically, the proposed enumeration method enables for the first time to measure the accuracy of the currently employed approximations. Our method is efficiently implemented with a few lines of codes, leverages parallel computations, and provably enumerates all the regions of the DN's partition. Lastly, our method has linear asymptotic complexity with respect to the number of regions and with respect to the DN's input space dimension. This property is crucial as we will demonstrate that sampling-based enumeration method has complexity growing exponentially with respect to the DN's input space dimension as a direct consequence of the curse of dimensionality \cite{bellman2015applied,koppen2000curse}. We hope that our method will serve as the baseline algorithm for any application requiring provable partition region enumeration, or to assess the theoretical findings obtain from the CPA formulation of DNs.

\vspace{-0.2cm}
\section{Enumeration of Single-Layer Partitions}
\label{sec:layer}
\vspace{-0.2cm}

We now develop the enumeration algorithm for a single DN layer. Because a DN recursively subdivides the per-layer partition, the single layer case will be enough to iteratively compute the partition of a multilayer DN  as shown in the next \cref{sec:deep}.

\vspace{-0.1cm}
\subsection{Layer Partitions and Hyperplane Arrangements}
\label{sec:MASO}
\vspace{-0.1cm}

We denote the single layer of a DN\footnote{without loss of generality we consider the first layer, although the exact same analysis applies to any layer in the DN when looking at the partition of its own input space} input-output mapping as $f_{\vtheta}:\mathbb{R}^D \mapsto \mathbb{R}^K$, with $\vtheta$ the parameters of the mapping. Without loss of generality, we consider vectors as inputs since when dealing with images, one can always flatten them into vectors and reparametrize the layer accordingly. The layer mapping takes the form
\begin{align}
    f_{\vtheta}(\vx)=\sigma(\vh(\vx))\text{ with } \vh(\vx)=\mW\vx+\vb\label{eq:layer}
\end{align}
where $\sigma$ is a pointwise activation function, $\mW$ is a weight matrix of dimensions $K \times D$, $\vb$ is a bias vector of length $K$, $\vh(\vx)$ denotes the {\em pre-activation map} and lastly $\vx$ is some input from $\mathbb{R}^{D}$. The layer parameters are thus $\vtheta \triangleq \{ \mW,\vb$. Although simple, \cref{eq:layer} encompasses most current DNs layers by specifying the correct structural constraints on the matrix $\mW$, e.g. to be circulant for a convolutional layer. The details on the layer mapping will not impact our results. The CPA view of DNs \cite{montufar2014number,balestriero2018spline} consists in expressing \cref{eq:layer} as
\begin{align}
    f_{\vtheta}(\vx) = \sum_{\omega \in \Omega}(\mA_{\omega}\vx+\vb_{\omega})1_{\{\vz \in \omega\}},\label{eq:CPA}
\end{align}
where $\Omega$ is the layer input space partition \cite{balestriero2019geometry}. Understanding the form of $\Omega$ will greatly help the development of the enumeration algorithm in \cref{sec:algorithm}. Given nonlinearities $\sigma$ such as (leaky-)ReLU or absolute value, it is direct to see that the layer stays linear for a region $\omega$ so that all the inputs within it have the same pre-activation signs. That is, a region is entirely and uniquely determined by those sign patterns
\begin{align*}
    f_{\vtheta}\text{ affine on $\omega$} 
    \hspace{-0.13cm}\iff\hspace{-0.13cm} \sign(\vh(\vx))\hspace{-0.1cm}=\hspace{-0.1cm}\sign(\vh(\vx')),\hspace{-0.07cm} \forall (\vx,\vx') \in \omega^2,
\end{align*}
where the equality is to be understood elementwise on all of the $K$ entries of the sign vectors. The only exception arises for degenerate weights $\mW$ which we do not consider since any arbitrarily small perturbation of such degeneracies remove those edge cases. From the above observation along, it becomes clear that the transition between different regions of $\Omega$ must occur when a pre-activation sign for some unit $k \in \{1,\dots,K\}$ changes, and because $\vh$ is nothing more but an affine mapping, this sign change for some unit $k$ can only occur when crossing the hyperplane
\begin{align}
    \sH_k \triangleq \{\vx \in \mathbb{R}^{D}:\langle \mW_{k,.},\vx\rangle +\vb_k=0 \}.\label{eq:hyperplane}
\end{align}
Leveraging \cref{eq:hyperplane} we obtain that $\partial \Omega$, the boundaries of the layer's partition, is an hyperplane arrangement as in $\partial \Omega = \bigcup_{k=1}^{K} \sH_k$.

We are now able to leverage this particular structure of the layer's partition to present an enumeration algorithm that will recursively search for all the regions $\omega \in \Omega$.

\vspace{-0.2cm}
\subsection{Region Enumeration Algorithm}
\label{sec:algorithm}
\vspace{-0.1cm}

From the previous understanding that the layer's partition arises from an hyperplane arrangements involving \cref{eq:hyperplane}, we are now able to leverage and adapt existing enumeration methods for such partitions to obtain all the regions $\omega \in \Omega$, form which it will become trivial to consider the multilayer case that we leave for the following \cref{sec:deep}.

Enumerating the regions of the layer $f_{\vtheta}$'s partition can be done efficiently by adapting existing reverse search algorithms \cite{avis1996reverse} optimized for hyperplane arrangements. In fact, a naive approach of enumerating all of the $2^K$ possible sign patterns $\vq \in \{-1,1\}^{K}$ and checking if each defines a non-empty region
\begin{align*}
    \bigcap_{k=1}^{K}\left\{\vx \in \mathbb{R}^{D}: \left( \langle \mW_{k,.},\vx\rangle +\vb_k\right)\vq_k \geq 0\right\}\overset{?}{=}\emptyset,
\end{align*}
would be largely wasteful. In fact, most of such sign combinations do produce empty regions e.g. if the partition is central i.e. the intersection of all the hyperplane is not empty then the total number of regions grows linearly with $K$ \cite{stanley2004introduction} and is thus much smaller than $2^K$. Instead, a much more efficient strategy is to only explore feasible sign patterns in a recursive tree-like structure. To do so, the algorithm recursively  sub-divides a parent region by the hyperplane of unit $k$. If that hyperplane does not intersect the current region then we can skip unit $k$ and recurse the sub-division of that same region by unit $k+1$. On the other hand, if hyperplane $k$ divides the current region, we consider both sides of it and keep the recursion going on both sides. We formally summarize the method in \cref{algo:police} and present one illustrative example and comparison against sampling-based region enumeration in \cref{fig:teaser}. In particular, we provide the efficiency of the sampling solution for various configurations in \cref{tab:times}.

\begin{algorithm}[t!]
\caption{\small
Proposed region enumeration method for the single hidden layer case that recursively searches over the feasible sign patterns $\vq$ one unit at a time, and only explores the branches that coincide with non-empty region i.e. avoiding the $2^K$ total number of possible of combinations. The step checking for intersection between an hyperplane and a given polytopal region can be done easily by setting up a linear program with dummy constant objective, the hyperplane as a linear constraint, and the polytopal region as inequality constraint; during the feasibility check the test will fail if the intersection is empty. This algorithm is obtained to provide the results from \cref{fig:teaser,tab:times}. The algorithm terminates once all the regions of the partition $\Omega$ have been visited.}\label{algo:police}
\begin{algorithmic}[1]
\Require $\mW \in \mathbb{R}^{K \times D}, \vb \in \mathbb{R}^{K},k\in\{1,\dots,K\},\vq\in\{-1,0,1\}^{k}$
    \State if ${\bf k\overset{?}{=}K+1}$ then this branch has reached a leaf, the sign pattern $\vq$ is feasible and can be accumulated into $\Omega$'s current estimate
    \State Check if the hyperplane defined by $(\vw_k,\vb_k)$ intersects the polytopal region defined by $\bigcap_{j=1}^{k-1}\{ \vx \in \mathbb{R}^{D}:(\langle \vw_j,\vx\rangle + \vb_j)\vq_j \geq 0\}$
    \State if {\bf NO} then unit $j$ is redundant, call the routine again with $[\vq_j,0]$ as $\vq$ and $k+1$ as $k$
    \State if {\bf YES} then unit $j$ splits the region into two, call the routine again with $[\vq_j,1]$ and $k+1$ and $[\vq_j,-1]$ and $k+1$
\Ensure $\mX^{(L)}$\Comment{Evaluate loss and back-propagate as usual}
\end{algorithmic}
\end{algorithm}

\begin{table}[t!]
\def\arraystretch{0.9}
\setlength\tabcolsep{0.9 pt}
    \small
    \caption{\small Comparison of our exact enumeration method versus sampling-based partition discovery for various single layer configurations with random weights and biases. The sampling-based discovery is run $5$ times and we report the average and standard deviation of the number of regions found after sampling. The number of input space sample is obtain so that the computation time of the proposed method is the same as the computation time of the sampling method i.e. for each configuration, both methods have run the exact same amount of time. We observe that for low-dimensional input space, and with the same fixed time-budget, both methods perform similarly and sampling is sufficient to quickly discover all of the layer's partition.}
    \label{tab:times}
    \centering
    % \begin{tabular}{c|l|l|l|l|l|l}\toprule
    %     width $K$                      & 16             & 32            & 64             & 128         & 256 \\ \toprule
    %     input dim. $D$                 & 2              & 2             & 2              & 2           & 2   \\ \midrule
    %     exact (ours).                  & 16             & 13            & 71             & 127         & 631 \\
    %     found by sampling              & 16 $\pm$0      & 13 $\pm$0     & 67$\pm$0       & 127$\pm$0   & 611 $\pm$2 \\
    %     $\%$ found by samp.         & 100            & 100           & 94             & 100         & 96 \\ \midrule
    %     input dim. $D$                 & 4              & 4             & 4              & 4           & 4   \\ \midrule
    %     exact (ours).                  & 54             & 80            & 1107           & 4271        & 95954 \\
    %     found by sampling              & 51 $\pm$0      & 69 $\pm$ 1    & 866$\pm$3      & 3288$\pm$18 & 70635 $\pm$55 \\
    %     $\%$ found by samp.         & 94             & 86            & 78             & 77          & 73 \\ \midrule
    %     input dim. $D$                 & 8              & 8             & 8              & 8           & 8   \\ \midrule
    %     exact (ours).                  & 24             & 1242          & 8396           & 386566        & - \\
    %     found by sampling              & 18 $\pm$0      & 543$\pm$2     & 2875$\pm$5      & 136748$\pm$251 & - \\
    %     $\%$ found by samp.         & 75             & 44            & 34             & 35          & - \\ \bottomrule
    % \end{tabular}
    \begin{tabular}{l|l|l|l|l|l|l}
        \multicolumn{2}{c|}{input dim \textbackslash width}                    & K=16             & K=32            & K=64             & K=128         & K=256 \\ \toprule
        \multirow{3}{*}{D=2}& enumeration                  & 16             & 13            & 71             & 127         & 631 \\
        &sampling              & 16 $\pm$0      & 13 $\pm$0     & 67$\pm$0       & 127$\pm$0   & 611 $\pm$2 \\
        &samp. found         & 100\%            & 100 \%          & 94 \%            & 100  \%       & 96 \%\\ \midrule
        \multirow{3}{*}{D=4}&enumeration                  & 54             & 80            & 1107           & 4271        & 95954 \\
        &sampling              & 51 $\pm$0      & 69 $\pm$ 1    & 866$\pm$3      & 3288$\pm$18 & 70635 $\pm$55 \\
        &samp. found         & 94  \%           & 86 \%           & 78  \%           & 77\%          & 73\% \\ \midrule
        \multirow{3}{*}{D=8}&enumeration                  & 24             & 1242          & 8396           & 386566        & - \\
        &sampling              & 18 $\pm$0      & 543$\pm$2     & 2875$\pm$5      & 136748$\pm$251 & - \\
        &samp. found         & 75  \%           & 44 \%           & 34 \%            & 35  \%        & - \\ 
    \end{tabular}
\end{table}

\begin{figure*}[t!]
    \centering
    \includegraphics[width=\linewidth]{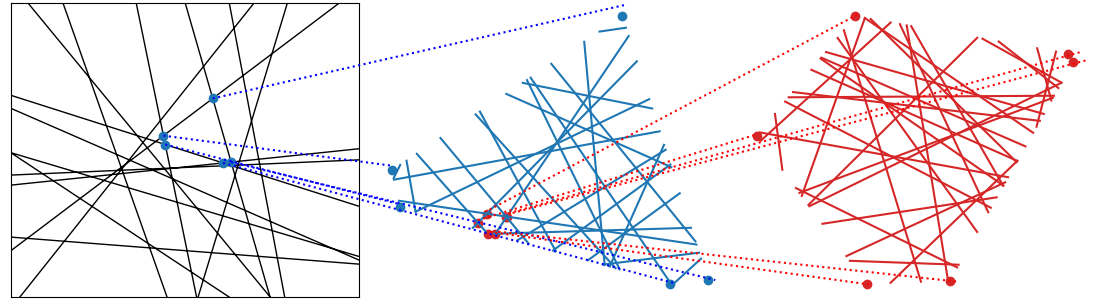}
    \vspace{-0.6cm}
    \caption{\small Depiction of the multilayer case which corresponds to a union of region-constrained hyperplane arrangements and thus which can be studied directly form the proposed hyperplane arrangement region enumeration. The only additional step is to first enforce that the search takes place on the restricted region of interest from the up-to-layer-$\ell$ input space partition. For example on the {\bf left column} one first obtains the first layer partition depicted in {\bf black}. On each of the enumerated region, a subdivision will be performed by the next layer; pick any region of interest, compose the per-region affine mapping (fixed on that region) with the second layer affine mappings, and repeat the region enumeration algorithm. This discovers the second subdivision done by the second layer, highlighted in \color{blue} {\bf blue }\color{black} in the {\bf middle column}. This can be repeated to obtain the subdivision of the third layer, here highlighted in \color{red} {\bf red }\color{black} in the {\bf right column}.}
    \label{fig:deep}
\end{figure*}

\vspace{-0.4cm}
\section{Enumeration of Multi-Layers Partitions}
\label{sec:deep}
\vspace{-0.2cm}

This section demonstrates how the derivation carried out in \cref{sec:layer} for the single layer setting is sufficient to enumerate the partition of a multilayer DN, thanks to the subdivision process under which the composition of many layers ultimately form the global DN's input space partition. We first recall this subdivision step in \cref{sec:subdivision} and summarize the enumeration algorithm in \cref{sec:deep_algo}.

\vspace{-0.2cm}
\subsection{Deep Networks are Continuous Piecewise Affine}
\label{sec:subdivision}
\vspace{-0.1cm}

We specialize the per-layer notations from \cref{sec:layer} by expliciting the layer index $\ell$ as $f^{(\ell)}$ for the layer mapping, as $\vtheta^{(\ell)}$ for its parameters, and the entire DN's input-output mapping is now referred to as $f_{\vtheta}:\mathbb{R}^D \mapsto \mathbb{R}^K$ with $K$ the output space dimension. The composition of layers take the form
\begin{align}
    f_{\vtheta}=\left(f_{\vtheta^{(L)}}^{(L)} \circ \dots \circ f_{\vtheta^{(1)}}^{(1)}\right),\label{eq:DNN}
\end{align}
where each layer mapping $f^{(\ell)}: \mathbb{R}^{D^{(\ell)}}\mapsto \mathbb{R}^{D^{(\ell+1)}}$ produces a {\em feature map}; with $D^{(1)}\triangleq D$ and $D^{(L)}\triangleq K$; with mapping given by \cref{eq:layer}, and $\vh^{(\ell)}$ denoting the {\em pre-activation map} of layer $\ell$. A key result from \cite{montufar2014number,balestriero2018spline} is the DN mapping is itself defined on a partition as in
\begin{align*}
    f_{\vtheta}(\vx) = \sum_{\omega \in \Omega}(\mA_{\omega}\vx+\vb_{\omega})1_{\{\vz \in \omega\}},
\end{align*}
which is known to be recursively built by each layer subdividing the previously built partition of the space \cite{balestriero2019geometry}.

\vspace{-0.2cm}
\subsection{Enumerating Union of Hyperplane Arrangements}
\label{sec:deep_algo}
\vspace{-0.1cm}

Considering an arbitrarily deep model can be tackled by understanding the recurrent subdivision process of a two hidden layer DN and applying the same principle successively. In this setting, notice that for the (two-layer) DN to be affine within some region $\omega$ of the DN's input space, each layer must stay affine as well. By composition the first layer staying linear does not ensure that the DN stays linear, but the first layer being nonlinear does imply that the entire DN is nonlinear. From that, we see that the first layer's partition are ``coarser'' the the entire DN's partition regions. More precisely, and following the derivation of \cite{balestriero2019geometry}, we obtain that each layer is a recursive subdivision of the previously build partition when in our case we need to search for each region $\omega$ of the first layer's partition the regions within it where the second layer stays linear. As a result, the proposed single hidden layer enumeration method from \cref{sec:layer} can be applied recursively as follows. First, compute the first layer partition enumeration. Then, for each enumerated region with corresponding sign pattern $\vq$, define a new single layer model with $\vh(\vx) \triangleq \sigma(\mW^{(2)}\diag(\vq)\mW^{(1)}\vx+\mW^{(2)}(\vq\odot\vb^{(1)})+\vb^{(2)}$ and within $\omega$ apply the single layer enumeration; repeating the process for all regions --and corresponding sign patterns $\vq$ of the previously found first layer partition. This enumerates the partition of $(f^{(2)} \circ f^{(1)})$, and the same process can be repeated as many times as there are layers in the DN; as illustrated in \cref{fig:deep}.

\vspace{-0.3cm}
\section{Conclusion and Future Work}
\label{sec:conclusion}
\vspace{-0.2cm}

In this paper, we provided the first exact enumeration method for Deep Networks partitions that relies on the existing highly efficient enumeration method of hyperplane arrangements. In fact, both the hallow and deep architectures produce partitions that correspond to hyperplane arrangements or union of restricted hyperplane arrangements. A crucial finding that was enabled by the proposed method is that sampling-based region enumeration, which is the only strategy used in current research studies dealing with DNs and affine splines, is in fact relatively poor at finding the regions of the DN's partition. In particular, when using such sampling to estimating some sensitive statistics e.g. the volume of the smallest region, sampling is biased and should be avoid in favor of an exact enumeration method.

\clearpage
\newpage

\bibliographystyle{IEEE}
\bibliography{bibliography}

\end{document}